# Dimensionality on Summarization

Hai Zhuge

*Key Lab of Intelligent Information Processing, Chinese Academy of Sciences, Beijing, P.R.China*
*Aston University, Birmingham, UK*

ABSTRACT

Summarization is one of the key features of human intelligence. It plays an important role in understanding and representation. With rapid and continual expansion of texts, pictures and videos in cyberspace, automatic summarization becomes more and more desirable. Text summarization has been studied for over half century, but it is still hard to automatically generate a satisfied summary. Traditional methods process texts empirically and neglect the fundamental characteristics and principles of language use and understanding. This paper summarizes previous text summarization approaches in a multi-dimensional classification space, introduces a multi-dimensional methodology for research and development, unveils the basic characteristics and principles of language use and understanding, investigates some fundamental mechanisms of summarization, studies the dimensions and forms of representations, and proposes a multi-dimensional evaluation mechanisms. Investigation extends to the incorporation of pictures into summary and to the summarization of videos, graphs and pictures, and then reaches a general summarization framework. Further, some basic behaviors of summarization are studied in the complex space consisting of cyberspace, physical space and social space. The basic viewpoints include: (1) a representation suitable for summarization should have a core, indicated by its intention and extension; (2) summarization is an open process of various interactions, involved in various explicit and implicit citations; and, (3) the form of summary is diverse and summarization carries out through multiple dimensions.

*Author's homepage:* http://www.knowledgegrid.net/~h.zhuge

# 1. Introduction

Versatile summaries accompany our daily life. Some summaries are mainly in form of text such as the abstracts of scientific papers, the prefaces of books, the tables of contents, CVs, the headlines of news, webpages with hyperlinks, book reviews, Wikipedia, and the results of Web search. Some summaries incorporate pictures, videos, graphs, or tables into texts. Applications include Web portals such as Yahoo, YouTube, posters, slides, medical certificate, TV guides, advertisements and conference programs. A good summary should be able to quickly attract attention, represent the core idea, and effectively convey the meaning according to interests. These summaries that people often see are made by humans.

1.1 *Automatic text summarization*

The development of cyberspace accelerates the expansion of texts since people can more and more easily and freely publish writings. Efficiently finding necessary contents in the ocean of texts is very important because life is short while new texts are continually generated.

With the development of sciences, researchers are limited in time and energy to read more and more publications. Researchers have to focus on the literature within recent years. This has led to more and more reinventions. Original innovation, especially systematic and fundamental innovation, has become more and more difficult.

Automatic summarization is a natural idea to solve this problem. Researchers have made great efforts to find the better solution. However, existing approaches are empirical and focus on special types of text. It is necessary to review previous efforts and explore the foundation of summarization.

There are different definitions of text summarization. The common point is regarding text summarization as an automatic process of distilling the most important language representation units from a text to produce an abridged version for a particular task and user [Mani and Maybury, 1999] [Mani, 2001].

The generic summarizers usually generate important contents in text(s) without considering users. The query-focused summarizers generate responses to user queries. The extractive summarizers select appropriate phrases or sentences from the text and then compose them. The abstractive summarizers can use different (probably more general) words to represent the main meaning of text. So far, most text summarizers are extractive. Important sentences can be extracted out according to the statistical analysis and experience on the input text while making abstraction needs knowledge on and beyond the text.

Automatic text summarization systems generally concern the following three issues:

(1) *Selection*. Scan and extract important language units (e.g., sentences).
(2) *Ordering*. Determine the order of the extracted units.
(3) *Realization*. Compose the extracted units to get fluent new text.

1.2 *Summarization of summarization literature: a multi-dimensional perspective*

Research on automatic text summarization started half century ago [Luhn et al., 1958] [Baxendale, 1958]. Research methods can be summarized by a multi-dimensional classification space [Zhuge, 2008, 2012], as shown in Figure 1.

Methods can be classified by input and output. There are three types of inputs: (1) *single text* [Luhn et al., 1958], (2) *multiple texts* [McKeown, 1995] [Radev, et al., 1998, 2002] [Nanba et al., 1999] [Agarwal et al., 2011], and (3) hybrid, which inputs one text and retrieves multiple relevant texts and then summarizes them so that readers can know more relevant contents. It is useful in summarizing a scientific topic [V.Qazvinian and D.R.Radev, 2008] [Chen and Zhuge, 2013].

There are two types of outputs: (1) *extractive summarization* (extracting important sentences from the given text) [Edmundson, 1969] [Mihalcea, 2005] [Shen, et al., 2007], and (2) *synthesized summarization* (summarization is a reformulated, compressed, abstracted, or synthesized text) [McKeown, et al., 1999] [Knight, 2002]. Sentences in output should be coherent to facilitate readers' understanding [Brandow et al, 1995] [Regina Barzilay and Mirella Lapata, 2008].

Methods can also be classified by techniques as follows:

(1) *Information fusion*. Summary is generated by identifying themes in text and selecting appropriate sentences for composition [Barzilay, 1999] [Barzilay & KcKeown, 2005].
(2) *Information retrieval*. Features such as frequency of words and phrases, locations, and ranks of sentences were used to extract important sentences [Baxendale, 1958] [Edmundson, 1969].
(3) *Machine learning*. Machine learning methods apply statistical techniques to extraction, including Bayes Methods [Kupiec et al. 1995] [Daumé and Marcu 2006][Louis, 2014], rich features and decision trees [Lin and Hovy, 1997], Markov Models [Conroy, 2001], Neural Networks [Nenkova, 2005], classification [Teufel 2002] [Pang and Lee, 2004], and hybrid machine learning method [Fattah, 2014].
(4) *Natural language analysis*. Methods based on natural language analysis were used in summarization [Barzilay, et al., 1997] [Silber and McCoy, 2002] [Erkan, 2004].
(5) *Classification and clustering*. Classification and clustering are the basic components of the multi-document summarization methods. It is usually used with graph analysis and information retrieval [Erkan, 2004] [Hilda Hardy, 2002]. Applications include summarizing positive and negative classifications in texts [Hu and Liu, 2004] [Pang and Lee, 2004].
(6) *Semantics-based*. Cognition scientists simulated human reading and understanding process as a series of propositions input and reduction cycles [Kintsch and Dijk, 1978; Britton and Graesser, 1996]. Text understanding is modeled by proposition network. The latent semantic analysis technique was used to identify semantically important sentences. Summarization patterns were discussed [Gong and Liu, 2001]. Discovering semantic community is a way to summarize a network of language units [Zhuge, 2009].
(7) *Other methods*, including *information extraction* (extracting entities, relations, and structures), *document compression*, *ranking method* [Rau, et al., 1989] [Daumé III, et al., 2004] [Carbonell, 1998], *probabilistic approaches* [Knight, 2002] [Qazvinian et al., 2010], and *citation-based approaches* [Abu-Jbara and Radev 2011] [Elkiss et al., 2008]. The faceted navigation approach is not traditional text summarization, but it is a special summarization because it can extract different facets in large text(s) to enable users to read only the interested facet [Xu and Zhuge, 2012]. Similarly, the association relation between words was used for multi-document summarization [O.Gross, et al, 2014]. The approach to summarizing differences between document groups was studied [Wang, et al., 2012].

Evaluation concerns human, semi-automatic and automatic methods based on the pre-defined standards [Mani and Maybury, 1999]. Methods for creating and evaluating summaries are usually coordinated each other [Hahn, 2000].

Every point in the space coordinates all methods specified at every dimensions.

Text summarization research has been extended to multi-medias and social events. Summarizing videos is a research topic in multimedia area [DeMenthon, et al., 1998] [Ekin, et al., 2004]. Some social events can be detected and summarized with wide use of online social networks [Zubiaga, 2012].

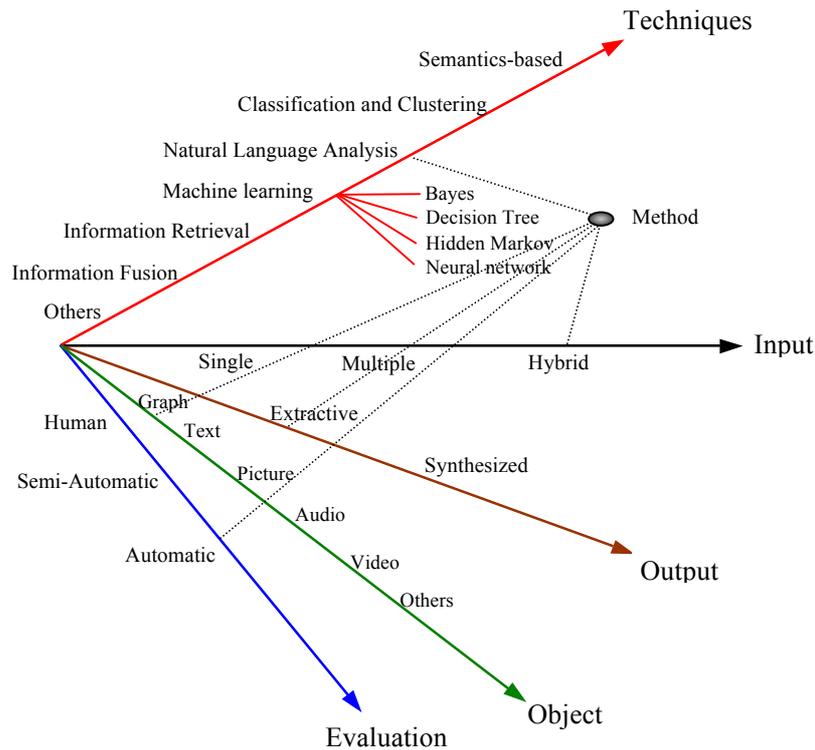

Figure. 1. A multi-dimensional classification space of summarization methods.

## 1.3 *Characteristics of text summarization*

Previous research methods have the following characteristics:

(1) *From text to text.* The main focus of research is on text itself. It is a natural idea to select and organize important sentences from the original text to form a summary. A different opinion is that language representations (e.g., words and sentences) indicate semantics rather than semantics itself, and interaction plays an important role in representation and understanding [Zhuge, 2010]. And, the knowledge for summarization is often beyond text. For example, 'A love story' is the high-level summarization of a novel, but the word 'story' may not be an important word in the novel or it even does not appear in the novel. Why can human use words beyond text? An explanation is that humans have commonsense on representation 'a novel tells a story' and representation 'story'.

(2) *Automation*. Summarization process excludes humans. It is ideal and efficient but summarization systems do not have minds and experiences as human. Humans experience with representation (at multiple levels), understanding and summarization in lifetime. Adding a ground with more indicators to a summary is a way to help representation and understanding, e.g., 'A love story in Qing Dynasty of China' includes a time indicator 'Qing Dynasty' and a location indicator 'China'. But, automatic summarization exclude interaction inevitably leads to an awkward summary.

(3) *Closed system*. The process of automatic summarization is closed, does not interact with other processes in cyberspace or social space. Actually, reviews and comments in cyberspace are open, easily available, and valuable for composing and improving summaries.

1.4 *Questions*

Many areas such as information retrieval and classification have significantly influenced research on summarization. Methodology has not been formed to guide research and development. It is critical for developing summarization research through thinking the following questions:

(1) What is summarization?
(2) Whether the best summary of a given representation exists or not?
(3) What are the fundamental principles and rules of language use and understanding behind summarization?
(4) What is the appropriate research methodology for studying summarization and developing summarization systems?
(5) Whether a general summarization exists or not?

1.5 *Summarization and dimensions*

Summarization is involved in representation and understanding. In language study, students are often requested to make summarizations after reading articles. Scientists summarize their ideas as abstracts placed before the main texts in papers so that readers can quickly know the main idea before reading the main text. Survey papers are summarizations of previous works on particular topics. Humans have the ability to represent and summarize what they have read. This ability can be enhanced through language learning, using and understanding, for example, journalists are specialized in summarizing events as news in the form that can attract readers. A text can be understood from different aspects because of the nature of language use and understanding.

Humans have the ability to summarize non-symbol representations, which can be regarded as a different dimension from text. For example, people can write summaries after attending conferences, watching movies, visiting museums and traveling in the physical space. Summarization enables people (from novice to experts) to quickly know the general and important information.

Humans live in a multi-dimensional space. The ability to understand, think through and use dimensions is an important part of human intelligence. The physical space including the nature and versatile artificial physical space like museum can be represented as a space of multiple dimensions such as time, region, and type (different types of museums may include different samples).

A set of representations can be classified by different methods. Regarding each method as a dimension forms a multi-dimensional classification space, where every point represents a class that has a projection at every dimension. For example, a publication space can include the following dimensions: *subject*, *time*, *author*, and *publisher*. The classification space can be normalized to ensure the effectiveness of operations on the space just like the normal forms of relational database. In a complex multi-dimensional classification space, a dimension can be a hierarchy of classifications and one point can be semantically linked to another [Zhuge, 2011, 2012]. Objects can be located in a space of multiple dimensions such as *time*, *topic*, and *publication type* (book, journal, or conference) for efficient retrieval and management.

A representation can be summarized from different dimensions.

## 2. Multi-Dimensional Methodology

There are different viewpoints on text understanding. For example, rationalism believes that the meaning of text is determined by its structure and derivation rules. The principles underlying the structure of language are biologically determined by human minds and genetically transmitted. Humans share the same underlying linguistic structure [Chomsky, 1986, 2006]. Social constructivism believes that any text is involved in society (e.g., in power relationships) and history [M. Foucault, 1966, 1969]. Evolutionism concerns the process of mental development and innate mental structures [Stern, 1985]. The innate mental structure that equips a man (especially a child) to interact with the world includes more than Chomsky's universal grammar of linguistic structure. The cyberspace, physical space and social space have structures, and the brains have evolved with ways to recognize and represent these structures and the structure of themselves. More phenomena have shown that what a man (especially a child) learns about the world is based on an innate mental structure [MacCarthy, 2007, 2008].

Any method is limited in its inventor's knowledge and understandings of problems. Integrating different methods is a way to break the limitations. Various methods can be organized in a multi-dimensional methodological space as shown in Figure 2. Each dimension consists of some specific methods.

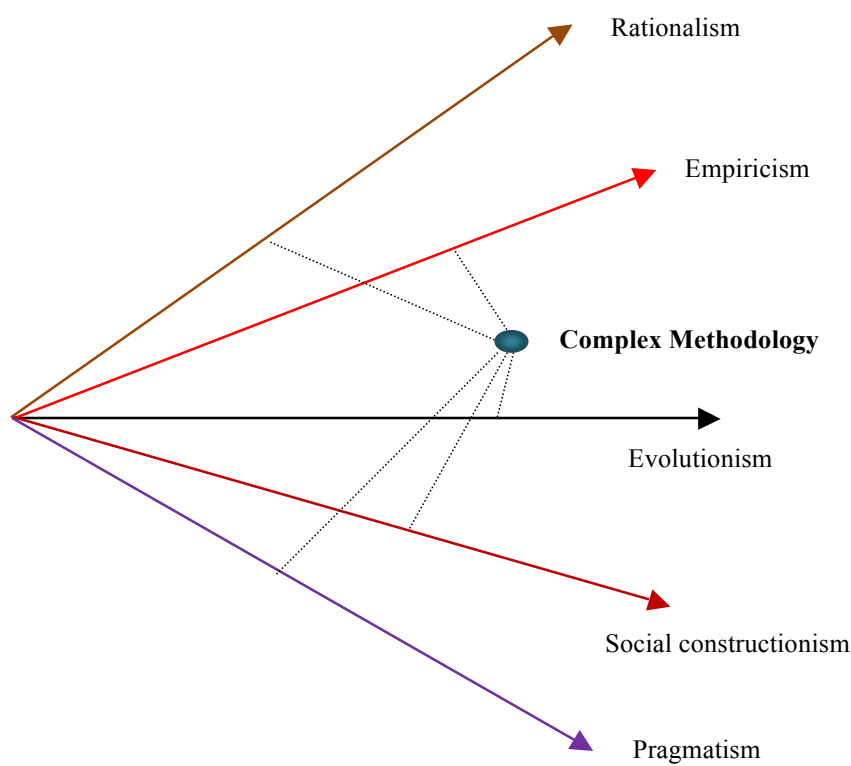

Figure 2. A multi-dimensional methodological space.

The space involves in the following high-level dimensions [Zhuge, 2012]:

(1) *Empiricism*. It believes that knowledge comes from experience and emphasizes evidence, especially data sensed through equipment or derived from experiment. It assumes that knowledge (including method) is convincible and reliable, at least within a certain scope.

(2) *Evolutionism*. It believes that complex systems or species develop through evolution. Summaries should be able to evolve with various interactions in the complex space like the evolution of the contents in Wikipedia. Writers, readers and languages evolve and influence each other.

(3) *Individual and social constructionism*. It believes that meaning and understanding develop with individual and society. For summarization, writing and reading involve in both individual thinking and social experience and interaction.

(4) *Rationalism*. It regards reasoning as the main source of knowledge. Rational study derives new theories and methods according to existing theories, methods and phenomena.

(5) *Pragmatism*. It argues that knowledge comes from practical use. Research should be useful and benefit human life and social development. Solving the problems emerged in practice provides instances for rational thinking and verification.

Different dimensions regulate different classes of method. Coordinating different dimensions is a way to generate new methods. A complex methodology is regulated by a subspace or a point, which has a projection at every dimension of the methodology space.

This paper attempts to explore the problem of summarization from multiple dimensions (especially from empiricism, rationalism, evolutionism, and social/individual constructionism) so as to form a general summarization methodology.

## 3. Basic Characteristics and Principles of Language Use and Understanding

As a kind of language representation, summarization should be based on the basic principles of language use. Observing and rethinking the basic characteristics of human language use can inspire research on summarization as language use including listening, speaking, reading, writing, understanding, and thinking.

**The basic characteristics of language use and understanding**

(1) *Human minds cannot directly access each other*.
(2) *Humans create and use languages to realize interaction between minds*.
(3) *Knowledge in minds evolves and self-organizes through language use*.
(4) *The use of language relies on knowledge*.
(5) *The representation of knowledge is not unique, and the understanding of representation is not unique*.

These characteristics lead to the following principles:

**Separation Principle**. The following three kinds of separations are involved in language:

(1) *Structure (grammar) and semantics are separated*. Semantics cannot be directly derived from structure. Actually, people do not rely on grammar to communicate with each other in daily life.
(2) *Knowledge for representation* (e.g., grammars and idioms) *and knowledge to be represented (e.g., scientific knowledge) are separated*.
(3) *Representation and its summarization are separated*. This leads to the separation of author's meaning, summarizer's understanding, and readers' understanding.

The above separation principles lead to the natural obstacle of summarization.

The problem of designing a suitable summarizer can be transformed into the problem of searching a suitable summarizer in cyberspace. The following are principles of selecting a suitable summarizer.

**Social Selection Principle**. *The suitable summarizer for summarizing a representation is rendered by the social network of its writers and readers*.

Knowledge selection is a kind of social selection.

**Knowledge Selection Principle**. *The person who shares the represented knowledge and the knowledge for representation is suitable for summarization*.

The above principle means that even people sharing the represented knowledge may not be the suitable person for summarizing the representation because they may not have the knowledge of representation. A friend of author may not be the suitable summarizer. The suitable person should have similar experience with the author (including education, work, etc.) through which knowledge is learned and shared.

The following lemmas can be derived from the knowledge selection principle:

**Suitable Summarizer**.

(1) *Authors who commonly cited a representation are the candidates of suitable summarizers*. This is because the represented knowledge was shared and the knowledge for representation is at the same level.

(2) *Authors are the best persons to summarize their own representations*. This is because authors have the knowledge for representation and the knowledge to be represented. However, authors have limited time and may be influenced by individual characteristics (e.g., health and mood, and worldview) and social characteristics (e.g., culture, economy and influence).

The evaluation of summarization follows some principles.

**Relativity Principle**. *Satisfactory of summary is relative. A summary that satisfies one person may not satisfy the other.*

Different persons can make different summaries for the same text, and one person can generate different summaries for the same text at different times. This is because knowledge of different people evolves personally. The inexact principle and the relativity principle indicate the following principles:

**Moderate Principle**. *A summary can be only moderately satisfied*.

The moderate principle indicates that the best summary does not exist. So, the pursuit of the best summary is insignificant.

**Dynamicity Principle**. *The satisfied summaries of a representation vary with time*.

This is because knowledge, interest and understanding are specific to people and change with the evolution of knowledge and society.

**Openness Principle**. *A satisfied summary can be reached only through an open social process of interactions and representations*.

The openness principle implies that establishing static criteria for evaluating summarization is unnecessary, and that a closed system is incapable for reaching a satisfied summary.

The above fundamental characteristics and principles indicate the following strategies for summarization.

**Summarization strategies**.

(1) *Making use of the summaries of the persons who have rich social relations with the original authors. More types of links render more common knowledge and experience* [Zhuge, 2009]. This is because the establishment of rich social relations indicates common individual characteristics and social characteristics.

(2) *Adapting to readers' interests*. The interests of readers determine the selection of summaries. A good summarizer should know its potential readers. This requests a summarizer to collect and analyze readers' interests according to their reading behaviors and attitudes to summaries. This is to pursue a suitable summary rather than the best summary.

(3) *Making summarization through human-machine interaction*, which can make full use of the advantages of both human and machines. It is the right way to pursue a satisfied summarization through a human-machine symbiotic system [Licklider, 1960].

(4) *Enabling different summaries to cooperate and compete with each other for impact at multiple dimensions* (e.g., acceptance for reading and adoption for generating new summaries). Ranking reviews to encourage contribution and competition among reviews reflect social value in a summarization environment. Different summaries may represent different characteristics of the summarizer. Integrating individuals of diverse characteristics can cooperate with each other in making new summaries.

(5) *Transforming summarizations into the problem of searching suitable persons or summaries in the social networks of authors, readers, summarizers, representations and links.*

(6) *Enabling summarizers to know the background of representations, including technological, social and economic aspects.*

## 4. General Citation — Definition, Measure and Axiom

Citation is the basic element of a scientific paper. It enables readers to trace the origin and access relevant knowledge. It reflects the author's thinking, comment, innovation, and summarization. It records the development track of science. Generally, citation is a kind of representation of selection and language use. Exploring citation is a way to explore the nature of summarization.

Summarization is requested at the advanced stage of language development when the complex structure of representation emerges. Citation is a basic semantic link of constructing a complex text. Therefore, it is necessary to understand citation when studying summarization.

Explicit citation is often used in scientific papers and books, in form of '[reference number]' or '(author, year)'. A research area emerges and evolves through continual citing a set of papers on the same set of concepts through time. As shown in Figure 3, a new paper *A* (denoting the title, author, abstract, etc.) becomes an often-cited paper and then becomes a source paper when the area is gradually formed. A survey paper summarizes an area through citing many papers in the area, and it is often cited as it helps later researchers to quickly know this area. During the development of a research area, different survey papers may appear at different development stages or on different facets [Afantenos, et al., 2005], the later survey paper can benefit from the summaries of previous survey papers. Citation and summarization are often involved in a reciprocity process [Novwak and Sigmund, 2005].

Different from static text, the citation network dynamically renders the source, the formation and evolution of the area, the backbone, the impact of researchers and institutions, potential knowledge flows through citation links [Zhuge, 2006], and the networks of cooperation between researchers and between institutions with the evolution of the area. Summaries of different scales can be obtained through zoom-in-and-zoom-out on the citation network. It is feasible to transform a citation network into a text by using some language patterns (for example, "the idea of A was extended by B", "the idea of A was used by B", and "the idea of A inspired B") to represent different citations, main roles, relations, and development track.

Hyperlink of the Web is a kind of explicit citation that freely complements, explains, or extends the content of the current Web pages. Homepage like Yahoo is the summarization of its web pages. Different from scientific papers, webpages can be changed, and links can be also changed, so the hyperlink network of webpages evolve notably. An advanced faceted navigator provides multi-facet summarization of the contents in a website for users [Xu and Zhuge, 2013].

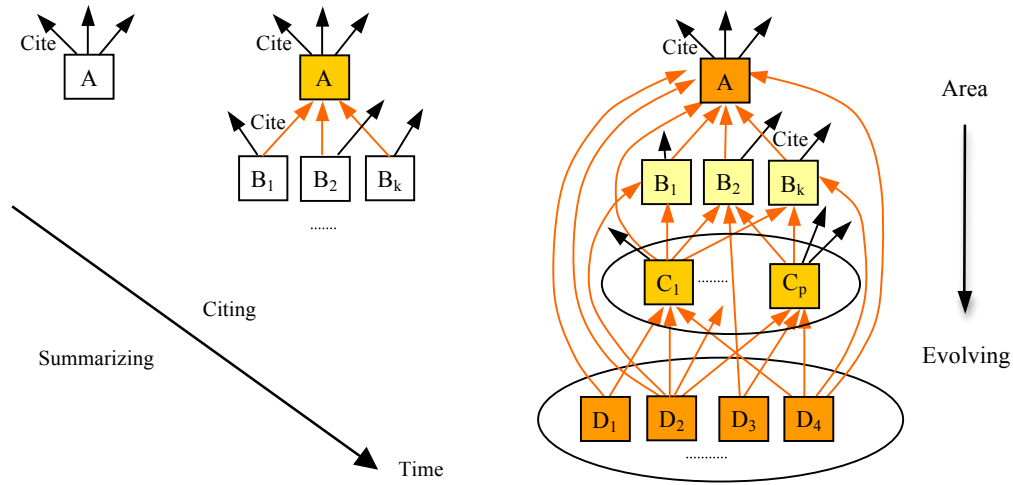

Figure 3. An area emerges when new papers often cite some old papers. The citation network evolves with partially summarizing an area and rendering topics, important roles, relations and development track.

Implicit citation is used in free texts and literature works. Some implicit citations have mark words such as "*someone says*" and according to "*someone's opinion*". These implicit citations can be located and transformed into explicit citation by searching these mark words and the references according to author names mentioned in text and then inserting uniform citation marks like scientific papers. Some implicit citations just reuse others' sentences or clauses without any mark word. For example, the sentence "It's Greek to me!" appeared in text implicitly cite the scenario used in 1599 in Shakespeare's play Julius Caesar, and "into thin air" cites the book written by Jon Krakauer published in 1997 and many other earlier works. Transforming implicit citations needs to compare common clauses in works published in different times.

Citation is a kind of representation by individual selection and language use. Summarizing various citations can reach a notion of general citation.

**Definition** (General Citation). *Citation is an individual selection of relevant representations for explaining, evidencing, complementing, commenting, or revising a representation, either explicitly or implicitly, according to individual motivation and knowledge. Diverse individual selections evolve the citation network*.

Citations form the intention and extension of a representation.

*The extension of representation* A consists of all representations that cite A and are cited by A. If representation A cites a set of representations B, then B constitutes an extension of A. If representation A cites a set of representations B, and A is cited by a set of representations C, then both B and C constitute the extension of A.

*The core representation* renders the core idea of a representation. A good article renders just one core. The core representation of a scientific paper is rendered by its keywords, title, abstract and conclusion. The representation that has a direct link to the core representation is ***close-core representation***. For example, two paragraphs sharing some words are linked by these common words. Representations linked to the close-core representations are relevant-to-core representation. Other representations are peripheral representations.

***The intention of representation*** *is indicated by core representations and by citation from other representations.* The intention of representation *p* is indicated by (1) *the core of p, Core*(*p*), consisting of the core representations of *p*; and, (2) *the representations that cite p*.

*A core representation has a high rank in the citation network.* It is usually emphasized in various ways to attract attention. For scientific papers, a core representation reflects motivation, problem or solution. The *intention* of a representation is rendered by commonsense, which is indicated by the basic representations. In natural language, commonsense is indicated by distinctive words and idioms. Some scientists tried to codify many commonsense to enable computers to have artificial intelligence beyond algorithm [Lenat, etc. 1991].

A core representation takes the priority of emerging when reading. In scientific papers, the core representations usually appear in the front and in the end (e.g., title, abstract and conclusion) so that readers can be impressed before and after reading the main body. This helps enhance the memory of the core by focusing and refocusing on the core when building or retrieving the semantic images in the mental space.

Humans have been composing complex representations and making summarization through times, so we have the following axiom.

**Axiom** (Additive Axiom). *A representation can be composed by a set of representations*.

This axiom is the basis of representation (including using languages) and summarization (especially, for multi-document summarization). Therefore, a representation *p* can be formalized as a structure of representations: *p*=*p* (*p*) | *p* ∪ … ∪ *p* | {*p*, …, *p*}, which represents a recursive structure of an abstraction *p* (*p*), an union *p* ∪ … ∪ *p*, and a set of representations {*p*, …, *p*}.

Figure 4 depicts the extension and the intention rendered by citations. Cite representation in scientific papers is rendered by the paragraph or the sentence that includes the cite mark commonly used in a community.

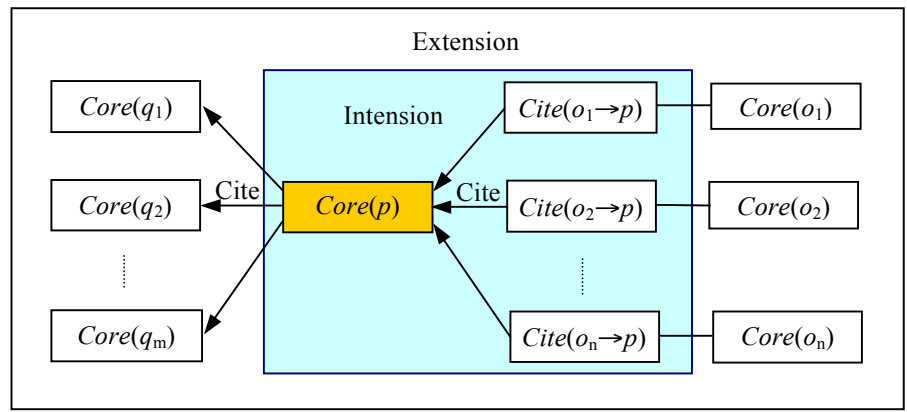

Figure 4. Extension and intention rendered by citation.

Dictionaries explain words in texts, so they can be regarded as the basic implicit citations to all texts. From this point of view, any text has a set of basic citations. So, the above statement is suitable for text.

*The basic behaviors of summarization include emerging, selecting, citing (explicitly or implicitly), and organizing representations according to requirement and motivation.* Text summarization is a special case of summarization.

**Definition** (**Summarization as citation**). *The summarization S of a set of representations P is rendered by its intention Int(P) and extension Ext(B) as follows, where S(Int(P)) is the summarization of the intention of P, S(Int(P)∪Ext(P)) is the summarization of the intention and extension of P, p→p' means that p cites p', {Cite($p_i$→p) | i∈[1, …, m]} denotes the set of cite representations in $p_i$ that cites p, and Cite($p_i$→p) describes p from the view of the author of $p_i$.*

$S(P) = <S(Int(P)), S(Int(P) \cup Ext(P))>$.

$Int(P) = <Core(p), Core(p) \cup \{Cite(o_k \rightarrow p) | k \in [1, …, n], p \in P\}>$.

$Ext(P) = \{Int(q_i) \cup Int(o_j) | p \rightarrow q_i, o_j \rightarrow p, i, j \in [1, …, n], p \in P\}$.

Different from previous notions of summarization, this definition gives the minimum summary (the summary of the intention) and the maximum summary (the summary of the intention and the extension) of a representation, and it regards citation as the fundamental behavior and mechanism of summarization.

## 5. Dimension of Representation

### 5.1 *Dimensions of structuring summary*

Structuring summary in an appropriate form is important as summary is mainly for human to read and understand. An appropriate form concerns the innovative cyber display based on human mental structure, psychological structure and innovative display. The following are some dimensions for organizing a summary.

(1) **Time**. *Organizing representations in time order or reverse time order.* Time order is in line with human innate sense of time and the process of reading.
(2) **Author**. *The original structure of the core representations takes the priority to appear in summary.* The reason is that authors are the best person to organize his representation. Authors can have their own styles (patterns) in organizing representations.
(3) **Core.** Features such as location, front size and color render a core representation. The core representations in one representation (e.g., text) render its topic. *Relevant representations are arranged at the places closed to the core representation.* This priority arranges relevant representations distributed in the original representation at the places close to the core representations in the summary. This is a phenomenon of semantic locality [Zhuge, 2010].

For a set of closely relevant representations, the order of organizing representations in a summary should consider *the formation process* of the set, which reflects certain semantics of the set. Citation between scientific papers reflects such an order, which should be considered in multi-document summarization.

The linear order of traditional text representations is in line with human physiological characteristics and innate sense of time. It is unclear so far how human mind organize knowledge. To reflect not only reading characteristics but also understanding characteristics, it is a reasonable method to combine the linear organization with the order of generalization and specialization.

Further, we can consider organize representations with a multi-dimensional classification space. A dimension like topic can organize coordinates as a tree representing multi-level generalization and specialization. Sub-dimensions can be arranged according to the measure of relevance between coordinates. Figure 5 shows a three-dimensional space for organizing representations through time dimension, author dimension and topic dimension. It enables readers to know the topic movement of a particular author or a group of authors through time. It also enables readers to know the role of author such as the source and the novice during the development of a topic at certain time. The multi-dimensional organization can provide multiple threads for readers to browse as indicated by the two-way arrows:

(1) Generalization and specialization threads through a topic tree.
(2) Time threads within a topic.
(3) Topic relevancy threads within a period at the time dimension.
(4) The evolution of topics in the area through the time dimension.

Humans experience in a multi-dimensional space but have to use a two-dimensional media such as paper and screen to externalize representation. Information loses through transformation from the internal representation to the form of display. Inventing a new interface that can easily convey representations through multiple channels is a way to improve human understanding.

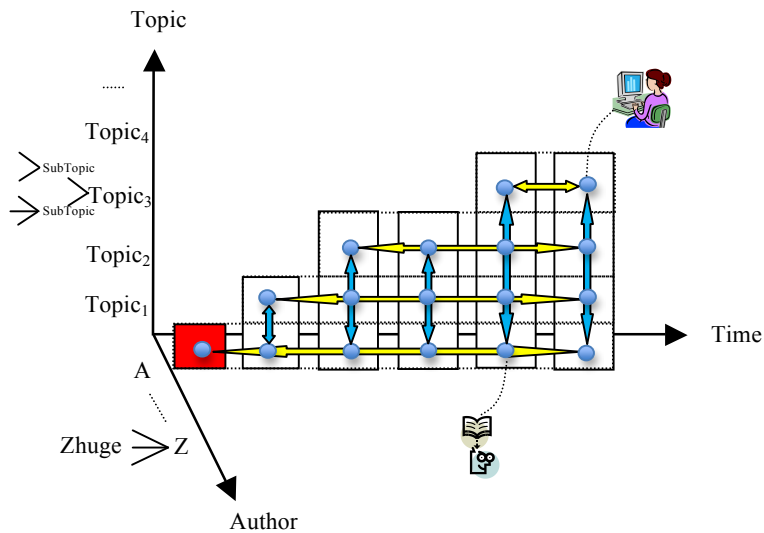

Figure 5. A three-dimensional space for organizing representations. The two-way arrows represent possible browse threads through the dimensions of time, topic and author.

A multi-dimensional space can be represented in different forms [Zhuge, 2008]. Figure 6 shows a space with four dimensions: *topic*, *region*, *time* and *author*, each of which is defined by a tree structure. Every point in the space has one projection at every dimension (a node in the tree). Moving from one point to the other point changes the projections at every dimension. This form of representation can represent a space of any number of dimensions. This form of representation looks similar to the wind-rose plot, which can efficiently convey the meaning of how wind speed and direction are typically distributed at a location. The difference is that the win-rose plot specifies the uniform value: speed, while a point in the multi-dimensional space can have different types of values (projections) at different dimensions.

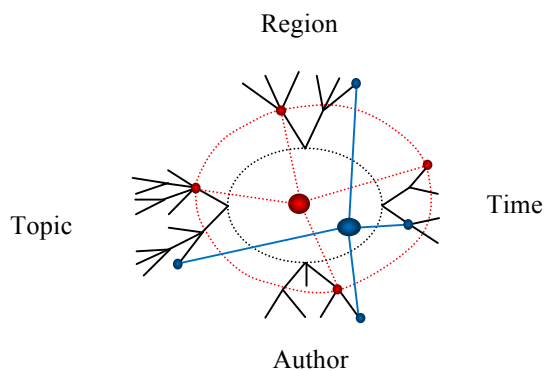

Figure 6. A display form of multi-dimensional space.

### 5.2 *Summarization on-demand*

Summarization carries out in an interaction environment where people read, write, cite and communicate with personal spaces. The personal spaces reflect personal reading experience, interests and knowledge based on the texts that have been read. A summarization can be satisfied only when it matches the personal space of the reader.

The general process of summarization on demand is shown in Figure 7. The summarization system is responsible for classifying, linking and reorganizing representations. The function 'Classifying & Linking' classifies representations according to the given dimensions, connects representations by discovering implicit relations, transforming implicit relations into explicit relations, discovering communities, and identifying appropriate representations. Users can adapt the dimensions to generate new summaries and to add summaries to the system for composition and comparison. Citation links between summaries and source representations help analysis and reuse. The information modeling provides the appropriate models for processing and organizing representations. The knowledge provides the rules of representation and understanding. The summarization strategies support the processing of representations under uncertain conditions.

The arrows in red color denote the following transformations:

(1) Transform representations (including, citation structure) into a multi-dimensional classification space (denoted as A in Figure 7) by classifying and linking representations.
(2) Transform a point or a subspace of *A* into a point or a subspace in the user personal spaces (denoted as *B*, *C* and *D*), which are also multi-dimensional classification spaces that represent users' interests and personalities.

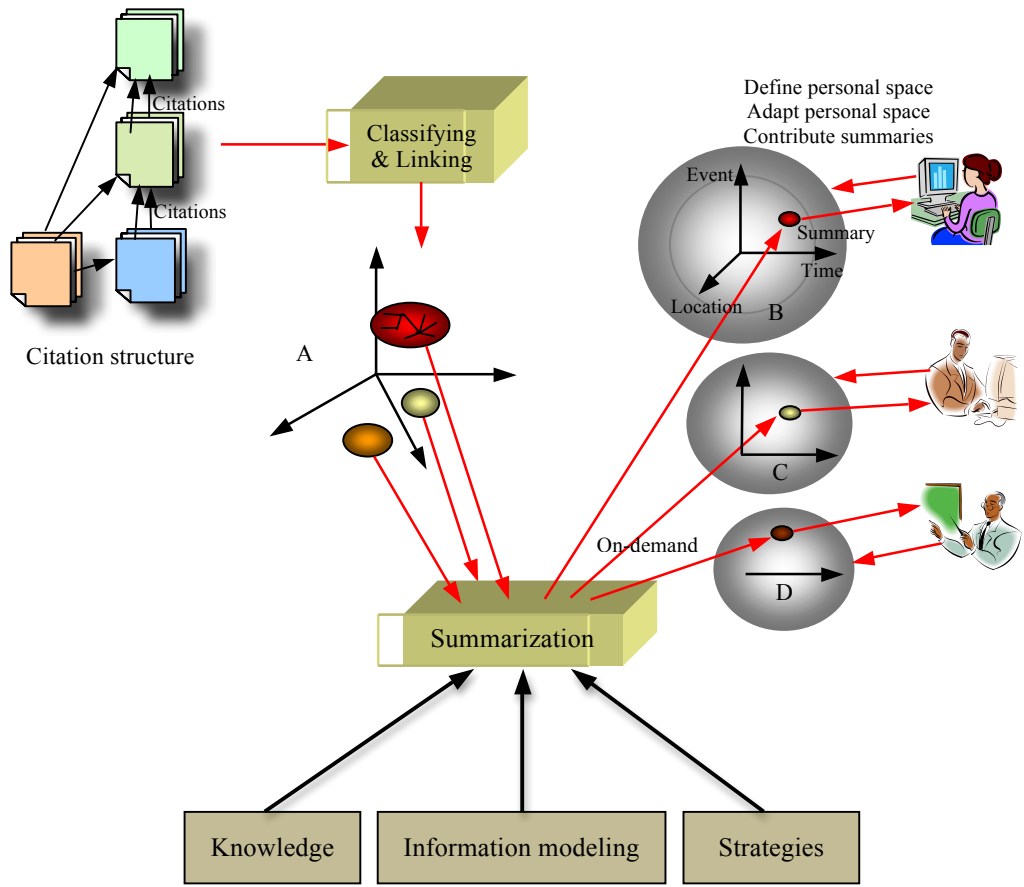

Figure 7. Summarization on demand.

## 5.3 *Forms*

Previous research on text summarization neglects the innovation of displaying summary as researchers assume that the form of output is the same as the input. However, the form of summary significantly influences the understandability of summary. Interface innovation enables users to easily understand a summary.

The following are possible forms of interface:

(1) *Simple structure*. Simple structures such as list, tree and grid have been widely used to summarize representations in daily life, e.g., displaying table of contents, family trees, organization structures, and properties on map. Different structures are suitable for different applications.
(2) *Hypertext*. A summary can be in form of a hypertext, where some texts are summarized as a hyperlink, and some texts containing hyperlinks are summarized as a hyperlink at the higher level. The advantage is that the content is the same as the original text and readers can read the concise top-level content first and then read the details by clicking the link if they are interested in. The following steps implement this:
(a) Give the expected size of summary and interest.
(b) Rank the paragraphs of the text according to the interest.

(c) Shrink the low rank paragraphs as hyperlinks.
(d) Do (a) if the size of the text is greater than expected, otherwise end. Automatic link generation was studied in hypertext area [Salton, et al, 1997]. An interactive visual text analysis tool can help readers understand the summary [Liu et al., 2012].

(3) *Semantic link network*. A graph with meaningful nodes and links can clearly summarize the main concepts and cues within text(s) [Zhuge, 2009, 2011]. A semantic link network of important concepts and relations can help readers quickly know the main cues and measures in representations. The key is how to extract appropriate nodes and links from representations. Different from linear reading, a semantic link network enables readers to know a general view and the main measures of large representations immediately. The semantic link network can be regarded as the map of the cyberspace and social space. Semantic link network can be regarded as the extension of the semantic net (Quillian, 1966) and correlational nets (Ceccato, 1961), which represents the relations between concepts.

(4) *Multimedia*. Coordinating texts, pictures and videos can render a summary from different channels of sense. It is particularly useful in making slides or posters for scientific research according to papers, and summarizing historical and literature works. A semantic link network is a way to organize multimedia.

(5) *Emerging*. This function is to emerge pictures or video clips relevant to the important representations in viewing scope while displaying. An eye tracking mechanism can help automatically locate the scope. Evaluation is needed to ensure the effectiveness of reading.

(6) *New devices*. New interface devices like 3D monitor will significantly influence the representation of summary. Optimizing the layout of display needs to consider three dimensions. 3D printer extends display from cyberspace to the physical space. 4D printer will further extend display to the formation of the objects sensed from more dimensions. The advanced summarizer could create new objects that can be seen, touched, smelled, heard and even tasted. People could hold what they think and write. The new interfaces provide a new cognitive environment for humans.

Enabling the form to accurately convey the core meaning of what is to be represented is the key.

## 6. Multi-Dimensional Evaluation

Previous evaluation methods focus on summary, which is just one dimension of summarization, and it is not reasonable to pursuit the best summary. The reasonable method is to study the multiple dimensions of the summarization environment and to consider the reasonable result.

Summarization is involved in individual and social construction processes, so it concerns multiple dimensions. The basic characteristics and principles of language use and understanding indicate the following dimensions for evaluating a summarization:

(1) **Reader**. It includes the following sub-dimensions:
   a) *Interest*. A summary should match the interests of readers. Sometimes, readers only need one aspect of a representation. Readers' comments and previous reading behaviors (e.g., the often clicked hyperlinks) reflect interests.
   b) *Cognitive level*. A summary should match the reader's cognitive level. Therefore, concepts at the reader's cognitive level should be selected.

(2) **Author**. A summarizer can understand the input representation better if it knows more about the author, including the relevant articles and social networks.

(3) **Input**. Input should also be evaluated because some representations are unnecessary for summarization. On one hand, a very simple and short representation such as a short para-graph of text and a picture is unnecessary for summarization as they can be quickly understood. On the other hand, the input should match readers' interest first.

(4) **Summarizer**.

   a) *Openness*. The closed systems designed by particular persons are unable to make satisfied summarization because only the persons sharing knowledge with the writer can make satisfied summarization.

   b) *Adaptability*. A summarization system should be improvable during use, and be able to adapt to readers' updates and adapt to new representations (especially on the same topic).

   c) *Interactivity*. This also implies that a summarizer should be able to interact with readers so that it can select the appropriate representations (e.g., sentences) and use the representations that the reader is familiar with. Further, it should interact with other people (including authors) to get more concepts and rules, and with the systems to get more information.

(5) **Output (Summary)**. It includes the following sub-dimensions:

   a) *Core representation*. The core representations and the relations between them should be reserved in summary.

   b) *Coherence*. Coherence between representations (e.g., sentences) can increase readablitiy [Brandow et al, 1995] [Regina Barzilay and Mirella Lapata, 2008].

   c) *Completeness*. A summary should be self-complete: all core representations that match reader's interest should be included.

(6) **Usage**. It includes the number of people who have used the summary and their attitudes.

## 7. Incorporating pictures into summary

It is a natural idea to incorporate pictures into summary since pictures have been widely used to render meaning in many representations. Many applications have incorporated pictures into summaries, e.g., transforming a paper into slides or a poster, transforming a novel into a carton book, and creating a webpage according to a set of texts and pictures.

There could be different ways to arrange pictures in displays but picture should be selected according to the core sentences and arranged near the core sentences according to the semantic locality principle [Zhuge, 2010].

The summary with pictures is more attractive than the text-only summary. A picture can convey meaning in about 1-10 seconds due to its familiarity and complexity to the viewer. In contrast, readers need to scan the whole text to know the text-only summary.

The reading order of the two summaries is different. Pictures take the priority in conveying meaning while reading the summary with pictures. The following are two kinds of reading order: (1) browse all the pictures → read the text beside each picture, and (2) view a picture → read beside text →···→ view last picture → read beside text. Pictures and texts may be viewed again but pictures still take priority during reviewing.

Further, the two types of summaries have different memory effects. The summary incorporating pictures can enhance reader's short-term and long-term memory. An explanation is that the summary with pictures gives readers stronger impression and provides more dimensions for rendering meaning, and establishes more links to render meaning.

A picture reflects a view of the physical space while natural language indicates a semantic image in mind. A summary rendered by both pictures and natural language provides more indicators for readers to build semantic images.

For extractive text summarization, to keep consistency between sentences can enhance readability. One advantage of incorporating pictures into summary is that pictures provide a different bridge between language units (e.g., sentences), especially when connection sentence cannot be found in original text. For example, the picture with tags *a* and *b* (e.g., *hotel* and *garden*, indicating "a picture of *hotel* with nice *garden*") can bridge the sentence containing *a* (e.g., *garden*, indicating "It is a beautiful *garden*"), and the sentence containing *b* (e.g., *hotel*, indicating "The *hotel* is near the sea"). This new bridge can enhance the readability of a summary.

New generation search engines have integrated pictures, the summaries of texts and hyperlinks in their formatted search results, which provides richer content for users than the link list provided by old search engines. However, a pre-designed format cannot adapt to user requirements.

The following are some problems and strategies.

(1) *How to select appropriate pictures?* Humans are specialized in recognizing pictures as they experience and reflect the physical space and form knowledge in mind. However, machines need to rely on human instruction to process pictures. The Web 2.0 provides the platform for people to upload and tag pictures on the Web. The *semantic link networks of texts, pictures, tags and users indicate a kind of social semantics of picture usage and thus provide the ground for selection*. So building the networks is the key to solve the problem. From the evolution construction and social construction point of views [Zhuge, 2012], tags might have been used in some texts by people. *Language representations like tags indicate the usage of the pictures*. Existing approaches such as feature-based approaches and machine learning approaches can be used to classify pictures. Image retrieval techniques can help find candidate pictures [Gudivada and Raghavan, 1995].

(2) *How to organize pictures and texts?*
   a) Use pictures to replace the corresponding representations in the original text, to summarize the rest representation, and to organize the summary according to the original structure.
   b) Select and use pictures to replace the texts in summary.
   c) Select and insert pictures into summary at appropriate places.
   d) Classifying pictures from multiple dimensions including time and location, which help distinguish pictures of different dimensions so that appropriate pictures can be selected to match the text in the summary.
   e) Construct a semantic link network of pictures, tags and language representations in relevant texts as the summary. A semantic link can be regarded as a citation that semantically connects two things [Zhuge, 2011].

(3) *How to identify events in pictures and link them to appropriate texts? The strategy is to make use of sensors and create semantic links between pictures and texts by detecting common projections at physical and social dimensions.* Current smart cameras (e.g., smart phones) can record the time and physical location of taking photos, which are the projections of pictures on the time dimension and the location dimension. The photos are probably relevant to the events happened at the same time and location.

Using pictures to summarize text is a new direction of summarizing text. Empirical research has been done in this direction [Zhu, 2007] [UzZaman, 2011] [Agrawal, 2011]. Research can lead to a new form of summary that can increase readability and understandability. Online picture-sharing systems like Flicker provide rich picture resources for implementing this idea. As new pictures are continually added to the online systems, a good summarization system should be able to keep up-to-date pictures in summarization.

Figure 8 depicts the idea of constructing a semantic link network of pictures and tags as a summary. The core words such as "CIKM2012", "hotel", "golf" and "garden" can be identified by comparing the source text and tags. Then, the relations like "back of" relying on the core words can be identified. So, the techniques of text summary can be extended to the construction of semantic link network and image retrieval [Gudivada and Raghavan, 1995]. Further, pictures can be extended to the snapshots of videos. Different from image retrieval (e.g., search images according to key-words), the picture-based summarization approach has a ground of texts (a network of source texts or summaries) when searching the picture-text repository.

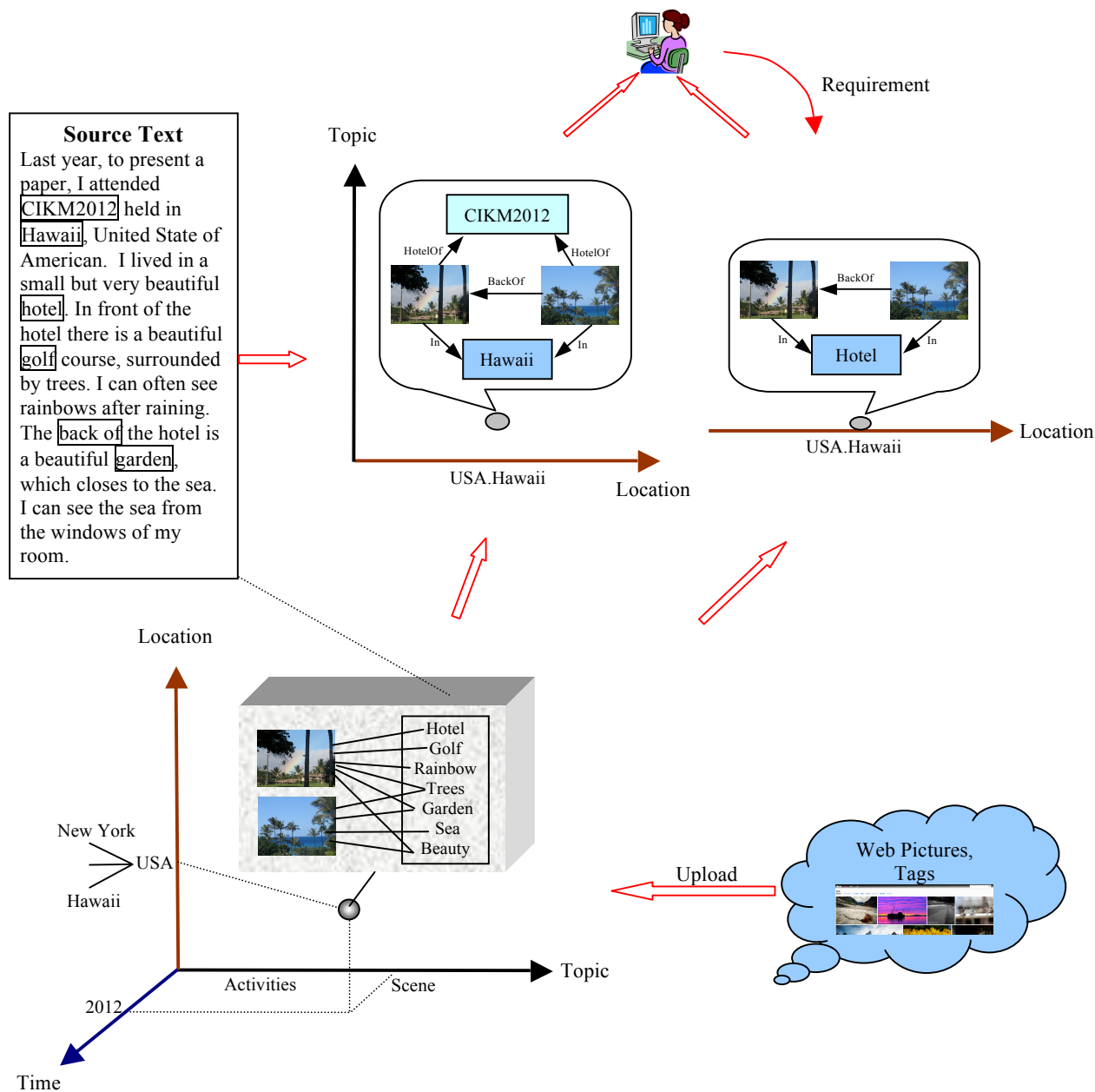

Figure 8. A summarization system consisting of a multi-dimensional classification space of summary in form of semantic link network of pictures and language representations and a requirement space defined and managed by users.

The generated summary is not unique as the tag sets of different pictures may be overlapped. Existing summaries can be put into the space and linked to the tag set. In this way, the existing summaries can be reused when making new summarizations.

It is important to ensure that the generated network of symbols and images should be small to facilitate understanding. According to the efficiency principle and regional principle, the radius of the network should be small. The radius can be defined as *the maximum length of link chain from the center (determined by core nodes) to any node*. In real application, a summarization system should enable readers to adjust the radius according to requirement.

Incorporating pictures into a summary enables a summarizer to summarize events. Events can be classified into points in a complex space with the following dimensions:

(1) *Time*. Dimensions evolve with time. Different types of events may have different distributions at the time dimension.
(2) *Location*. An event happens at a physical location, which can be captured by GPS, IP, or communication network.
(3) *People*. Different classes of people play different roles in society and thus are likely involved in different events.
(4) *Category*. The category hierarchy of events.
(5) *Representation*. It usually includes some sub-dimensions: *feature representation*, *pattern representation*, *language representation* (including text, voice, and movie), and *function representation* (different objects such as car, mobile phone, house and road have different functions)

## 8. Summarizing Videos, Graphs and Pictures

8.1 *Summarizing videos*

Automatic video summarization is to enable machines to generate a clip of a long video or a set of long videos. It is important in video management, retrieval and browsing. It becomes more and more important with wide use of digital cameras in our society for security, news, entertainment, education, advertisements, etc. Humans are able to make operations to summarize a movie according to their understandings and requirements, but are limited in ability to view and summarize huge volumes of videos generated everyday. Automatic video summarization can help humans quickly know the key content in big video volumes.

There are two fundamental classes of video summarization: (1) focusing on still picture (static storyboard), a small collection of salient images extracted from video sources; and, (2) focusing on moving pictures (moving storyboard), a collection of image sequences, and the corresponding audio abstract extracted from the original sequence, which results in a short video clip.

Summarizing a video requires machines to identify the essential characteristics of the video. Video summarization concerns the simplification of motion. A video sequence can be represented as trajectory curves in a high-dimensional feature space, which can be decomposed into curve segments of low dimension for simplification [DeMenthon, et al., 1998]. The patterns of events, conversations, and behaviors should be characterized to get semantically meaningful summaries of complicated video contents [Tewfik, 1999]. Low-level features such as color, boundary, and shot classification can help summarize videos. For particular applications like football games [Ekin, 2003], important sections such as slow-motion segments and goals in game are known by humans, so it is easy to find the important sections. Some criteria such as *coverage* (the summary should represent the original one) and *diversity* (the elements of the summary should be distinct from each other) were proposed [Shroff, 2010]. Some approaches focus on some contents about who, what, where, and when in the framework of the video contents to produce a concept-level summary [Chen, 2009]. Existing research is generally empirical and focuses on particular applications.

Modern movies provide more channels (e.g., voice and music) for understanding than early silent movies. Current online movies contain subtitles, which provide a new condition for making summarization of videos through natural language processing. These subtitles provide the basis for generating a piece of text as the summary of movies.

A semantics-based approach to summarizing videos is to construct a semantic structure on videos by introducing semantic links into videos. The higher abstraction level of the structure presents more general summary of the video. A semantic link network of video components enables users to query the interested components and navigates in the network according to interest, e.g., the semantic link network of video components can help play the main thread of a story development.

To realize more meaningful video summarization, it is important to represent body language, spoken language, emotion, habit and psychological activities in microscopic. Behavior recognition is the basic components of understanding videos. In macroscopic, it is important to represent the background of the video and relevant social structure, interests, fashion, rules, regulations, laws and culture of society in a summarization system. These concern in-depth understanding of the basic interaction principles in the complex space consisting of cyberspace, social space and physical space.

8.2 *Summarizing graphs*

Graphs are the generalization of various real networks in cyberspace (e.g., data structures), physical space (e.g., supply chains, material flow networks) and social space (e.g., social networks of things).

Summarizing data structures can provide more general data services. The summarization of relational database takes table as input and produces a reduced version of the table through rewriting and generalization. The resulting table is expected to provide tuples with less precision than the original but more informative [Saint-Paul, 2005]. Summarizing data streams supports more general fundamental queries on data streams such as point, range, and inner-product queries [G.Cormode and S. Muthukrishnan, 2005]. Fuzzy set was used to summarize data structures [Yager, 1982] [G. Raschia and N. Mouaddib, 2002]. The Resource Space Model supports multi-dimensional generalization and specialization [Zhuge, 2008, 2012]. Transformation between different data structures can help summarization [Zhuge, 2008]. A unified representation such as XML (Extensible Markup Language) and OWL (Web Ontology Language) helps transformation and unify summarization.

Summarization can help humans understand large-scale graphs. Summarizing a large-scale graph of data is important to graph data management as it can render the patterns hidden in data [Navlakha, 2008] [Tian, 2008]. An interactive graph summarization approach was proposed [Zhang, 2010]. Statistics is a useful means for machines to summarize graphs. Network analysis techniques such as degree distribution and community discovery can be used to find more important part and the hierarchy of a graph.

In some areas like CAD, graph components have formal specifications. Making abstraction on graph components can be done through mathematical derivation. So, summarization in these areas has the reasons of summarization and the correctness guarantee of using summaries.

If we regard a text as a graph of words or sentences, text summarization can be regarded as a problem of summarizing a semantic link network [Zhuge, 2009], where nodes and edges can be texts, pictures and videos.

8.3 *Summarizing pictures*

The main purpose of summarizing pictures is to generate a small set of pictures from a large set of pictures according to interest. There have been more and more real requirements of summarizing pictures with the explosion of digital pictures online in recently years due to the wide deployment of cameras and popularity of smart phones. The summarization of pictures also helps incorporate appropriate pictures into the summary as discussed before.

Humans can make a humanized summarization because they have experience and knowledge out of pictures. It is hard to enable machines to make a humanized summarization. Discovering the se-mantic link networks that the pictures involved in can help automatic summarization. Automatic summarization of pictures can be extended to include the solutions to the following issues:

(1) *Generating a piece of text to represent a set of pictures*. A solution is to transform this issue into an information retrieval and text summarization issue: Select the representative tags of these pictures, search relevant texts according to these tags (or select the texts that contain or link to these pictures), and then summarize these texts. Another way is to establish basic semantic links between pictures, find the best matched text, and make necessary text summarization. The establishment of the semantic links relies on the relations between their tags determined by existing texts and links and the categories of pictures.

(2) *Selecting one picture to represent a set of related pictures*. A solution is to transform this issue into a text summarization issue: Transform pictures into texts according to the way described in (1), summarize the texts, and then select a picture to represent the summary by matching its tags and the core words of the summary.

(3) *Generating a small network of pictures from a large set of related pictures*. A solution is to establish the semantic links between pictures, discover the communities of the semantic link network of pictures, select one picture to represent one community, and construct a network of the representative pictures.

(4) *Generating a small network of texts according to a large set of pictures*. A solution is to discover the communities of the semantic link network of pictures, to select one text to represent one community according to the tags of pictures within the community, and to construct a semantic link network of the representative texts.

A solution to implement this idea is to make use of existing summaries made by humans and the corresponding pictures in the networks of pictures, tags, summaries, source texts and people who involved in forming and using these things.

The key problem is to select a better picture from the candidates that have the same projections on the dimensions of time, location and topic, because the semantic links between pictures may be poor. For example, it is hard to automatically find the abstraction relationship between pictures.

Modern cameras can generate pictures with time and location information, so the summarization of pictures can be carried out in a space of four dimensions: *language*, *feature*, *time* and *location*.

The problem of summarizing texts, pictures, and videos can be generalized as one problem. The relations between texts, between pictures, and between videos can be mutually enforced, explained and rendered. The form of summary can be a semantic link network of texts, pictures, audios, and videos.

## 9. General Summarization

9.1 *Unification*

Humans have been pursuing the ways to represent thoughts, behaviors, artifacts and the nature. Various devices and approaches have been invented and developed to represent and process different forms such as natural languages, pictures, videos and graphs. Various representations constitute a representation space with particular structure and operations. Summarization is a kind of operation that inputs one or more representations in this space and then outputs a new representation.

Let's recall how human process representations generated through different channels. People have the following common experience: Scanning the symbols in a novel generates some images in mind, and the images emerge before symbols when recall. Similar images will be generated when seeing the movie about the novel. A distinguished characteristic is that humans generate behaviors (including mental behaviors) different from the input. This indicates that *human mind uniformly* *processes various things at certain cognitive level*. This indicates the possibility of creating a unified method for processing different objects such as texts, pictures, videos and graphs.

Cognitive psychologists argue that people usually remember meaning rather than exact representation and that meaning represents through the perceptual and motor systems for interacting with the world. The categorical organization of knowledge strongly influences the way to encode and remember experiences [Anderson, 2010]. This is the psychological basis of classification.

Humans represent what they have seen or felt as semantic images in the mental space through interacting and experiencing in the physical space, and summarizing representations and revising representations during communicating with each other in social space where motivations are generated. To facilitate communication, humans indicate the mental semantic images in the language commonly used in society, but it is hard to communicate in the form of what they have seen or felt. A semantic image can be an image-like form when representing daily life or a symbol-like form when representing abstract concepts.

Discovering unity in diversity is a scientific research method, which has generated many important scientific principles and theories, for example, Maxwell successfully unified electricity and magnetism. However, unifying different theories is hard because it needs to uncover the common nature behind existing theories, and different theories may represent different aspects of a domain and use different representation systems that are difficult to be unified. Sometimes, pursuing unity is an adventure like the pursuit of a unified field theory.

Knowledge representation approaches such as the production rule [Davis, et al, 1977], the frame [Minsky, 1975] and the semantic net [Quillian, 1966] are symbol systems that can carry out reasoning for solving problems. A unified representation should reflect the most fundamental characteristics of concerned representations. The Unified Modeling Language (UML) uniformly represented business processes and behaviors, software architectures, processes and behaviors, and data structures.

Establishing a unified representation enables a summarization system to uniformly process various representations. A transformational development of the fundamental infrastructure of cyberspace (e.g., a new generation computer) and smarter devices will influence the generation of new representations.

## 9.2 Transformation with dimension reduction

A text can be transformed into a semantic link network of words, sentences and paragraphs. One semantic link network of texts can link to the other semantic link networks of texts to form a larger semantic link network through such relations as citation and coauthor.

A video can also be transformed into a semantic link network of video clips. Links between words in subtitles and video clips can be established. Links between videos are enriched through the mediation of words. For the videos with scripts, words will play more important role in representing videos. In addition, voice and music can render the link between videos.

A unified method for summarizing different forms can be developed by transforming texts, videos, pictures and graphs into semantic link networks. A complex semantic link network of different forms enables one form (e.g., movies) to link to other forms (e.g., novel, script, pictures of actors and actress, posters, comments, related movie, etc). The form of summary can be a small semantic link network of texts, videos, audios, pictures and graphs, which provides more semantic indicators than single type of form like silent movie. Appropriate coordination between different forms concerns humanity and sociology. The size of summary depends on user requirement and cognitive level.

Cyberspace consists of huge links among texts, videos, audios, pictures and graphs, corresponding to human senses and the structures of cyberspace, physical space and social space. Any text, video, audio, picture or graph does not exist independently, has explicit or implicit links to other form of representations. One form of representation like text usually links to other forms of representation like pictures, and possibly to video, audio and graph, which can be regarded as citation for explanation, complementation or extension. The network evolved with social interactions determines different summaries at different times.

Transformation between representations is a way to realize unification. It inputs one form of representation and outputs another form of representation. Summarization is a special transformation that operates dimension reduction for easier understanding. From this point of view, summarization can be regarded as a transformation of reducing the dimensions of a representation so that the dimensions of representation can be linked to and merge with the dimensions in the mental space. Therefore, we have the following definition.

**Definition (Multi-dimensional summarization).** Multi-dimensional summarization is a function $S(P(d_1, ..., d_m)) = T(d_i, ..., d_j)$, which transforms a representation $p(d_1=p_1 ..., d_m=p_m)$ in the source space of representations *P* with dimensions $d_1, ...,$ and $d_m$ into a representation $t(d_i=t_i, ..., d_j=p_j)$ in the target space of representations *T* with dimensions $d_i, ...,$ and $d_j$ ($m ≥ j ≥ i ≥ 1$) such that *t* contains the core of *p* at dimensions $d_i, ...,$ and $d_j$.

The dimensions of *P* vary with different sources and the dimensions of *T* vary with the readers of the summary. Some relations may exist between dimensions. Usually, *T* has a small number of dimensions. The scale of summary can be regarded as a dimension. The reason is that a reader can easily and quickly understand the summary if the dimension of the target space is the same as the dimension of the reader's personal space, which represents the reader's mind (cognitive architecture).

The definition of summarization based on citation given in section 4 gives the range of *T*, and $(d_i, ..., d_j)$ reflects the basic cognitive level of reader.

9.3 *Cognitive level*

Cognitive psychologists have been exploring mental concepts through rational definition, prototype, exemplar and knowledge studies. They try to find the basic cognitive level in the concept hierarchy shared by people [Murphy, 2002]. However, different communities can have different cognitive levels. The basic cognitive level can be established for the hierarchy of universal concepts. The cyberspace including the Wikipedia is reflecting more and more of the hierarchy of the universal concepts. The cognitive hierarchy of different communities corresponds to different sub-graphs of the hierarchy and has different basic cognitive levels. The cognitive level for a particular research field can be reflected by all of its papers. It stands for the basic cognitive level of all the authors in the field.

Relevant research concerns commonsense, knowledge level, and abstraction [Newell, 1982] [Minsky, 2006] [Tenenbaum, 2011]. Physical instruments such as functional magnetic resonance imaging fMRI and electroencephalography EGG have been used to detect the physical status of mind [Turkeltaub, 2002]. The relations between language and brain have attracted many researchers [Friederici, 2000] [Bates and Dick, 2002].

*A summary is suitable if its cognitive level is the same as the reader's cognitive level.*

The following are some rules to make a suitable summary:

(1) *If the cognitive level of the original representation is the same as the reader's cognitive level, the summary should use the core representations in the original representation.*

(2) *If the cognitive level of the original representation is higher than the reader's cognitive level, the summary should use more specific concepts in the commonsense category hierarchy.*

(3) *If the cognitive level of the original representation is lower than the reader's cognitive level, the summary should use more general concepts in the commonsense category hierarchy.*

9.4 *Representation lattice*

In psychology, representation is a kind of hypothetical internal cognitive symbol that represents external reality. Externalization of the internal representation involves in complex mental, physical and social behaviors. Ontology helps establish a general representation from the nature of the world, knowledge and knowing. From the pragmatism point of view, ontology was developed by IT professionals for information sharing [Gruber, 1993] [Ashburner, 2000]. Ontology helps explain representation and establish the links between representations. This enables summarization systems to use more general or specific concepts in summary.

Representations can be generalized, united, classified and semantically linked to form a lattice of representations at a cognitive level of ontology as shown in Figure 9. A cognitive level determines a representation lattice. Operations on representations enrich the structure of the lattice. Abstraction and analogy are the important operations of generating representations [Zhuge, 2010, 2011, 2012].

The abstract representation reflects the common characteristics of a set of different types of concerned representations. Abstract representation is particularly useful for developing theories. Mathematical tools such as logic, algebra and graph can help develop abstract representations. However over abstract representation may not be useful in real applications.

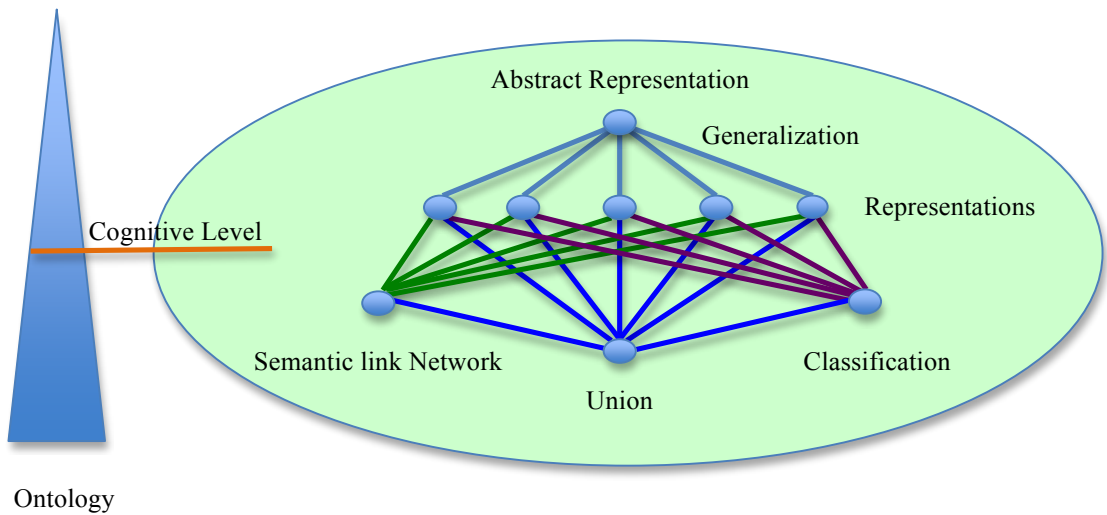

Figure 9. Representation lattice at certain cognitive level.

The union of representations integrates different representations to provide a global view of these representations. It is important to ensure the understandability and expressiveness of the integrated representation. As a kind of union, data integration enables users to get a global view of data generated from different sources [Friedman, et al, 1999; Lenzerini, 2002; Halevy et al, 2006]. The union of the semantic link network and the classification space forms a complex classification space. A complex space incorporating multi-dimensional classification space and semantic link network was used to represent and organize semantic images. Different representations of the same thing can be linked to the same semantic image for understanding and thinking [Zhuge, 2010, 2011, 2012].

The general summarization inputs a representation and a cognitive level at certain ontology and then outputs a representation lattice and a recommended representation.

## 12. Conclusion

Summarization is an open representation of representation, in diverse forms, from multiple dimensions, and through interactions in multiple spaces. The basic interactions include selecting, citing and organizing representations according to requirement and motivation.

Texts have been the major means to reflect the evolution of human society. It is necessary to explore automatic summarization of texts expanding rapidly in cyberspace. However, existing approaches are empirical, mainly relying on statistics, structure and linguistic analysis while neglecting the nature of representation and understanding. This paper summarizes previous approaches and explores the fundamental theory and method for summarization from multiple dimensions. The core viewpoint is that summarization is carried out with various interactions involved in human, machine, and various representations including text, picture, and video. Writers and readers represent mainly through explicit or implicit citations. Studying the summarization of pictures, videos and graphs reaches a general summarization method.

It is hard for automatic summarization systems to realize human-level summarization as it concerns the essential natural and social differences between human and the systems. Putting summarization into a human-machine-nature symbiotic network is a way to make a breakthrough. A significant progress of summarization research relies on an innovative summarization of philosophy, psychology, linguistics, cognitive science, neuroscience, physics, computer science, and artificial intelligence from multiple dimensions.

This work is an attempt to help broaden the scope of summarization research and inspire cross-disciplinary research to develop this area.